\pdfoutput=1

\documentclass[11pt]{article}

\usepackage{emnlp2021}

\usepackage{times}
\usepackage{latexsym}
\usepackage{soul}

\usepackage[T1]{fontenc}

\usepackage[utf8]{inputenc}

\usepackage{microtype}

\usepackage{epsfig}
\usepackage{amsmath}
\usepackage{amssymb}
\usepackage{microtype}
\usepackage{xspace}
\usepackage{subfig}
\usepackage{graphicx,wrapfig}
\usepackage{booktabs}
\usepackage{makecell}
\usepackage{framed}
\usepackage{multirow}
\usepackage[linesnumbered,algoruled,boxed,noend]{algorithm2e}
\usepackage{enumitem}

\newcolumntype{L}[1]{>{\raggedright\let\newline\\\arraybackslash\hspace{0pt}}m{#1}}
\newcolumntype{C}[1]{>{\centering\let\newline\\\arraybackslash\hspace{0pt}}m{#1}}
\newcolumntype{R}[1]{>{\raggedleft\let\newline\\\arraybackslash\hspace{0pt}}m{#1}}

\SetKwInput{KwInput}{Input}
\SetKwInput{KwOutput}{Output}
\SetEndCharOfAlgoLine{}

\newcommand{\para}[1]{\smallskip\noindent\textbf{#1}}
\newcommand{\txt}{t}
\newcommand{\img}{v}
\newcommand{\sal}{S}
\newcommand{\mat}[1]{{{\mathbf{#1}}}}

\DeclareMathOperator*{\argmin}{arg\,min}

\usepackage[tikz]{bclogo}
\definecolor{msftBlue}{RGB}{0,164,239}
\definecolor{msftGreen}{RGB}{127,186,0}
\definecolor{msftYello}{RGB}{255,185,0}
\definecolor{msftBlack}{RGB}{0,0,0}

\begin{document}

\title{MSD: Saliency-aware Knowledge Distillation \\for Multimodal Understanding}

\author{
Woojeong Jin$^{1}$\thanks{\xspace\xspace The work in progress was mainly done during internship at Facebook AI.}  \quad Maziar Sanjabi$^2$ \quad Shaoliang Nie$^2$ \quad Liang Tan$^2$ \quad Xiang Ren$^1$ \quad Hamed Firooz$^2$\\
$^1$University of Southern California \quad $^2$Facebook AI\\
{\tt \{woojeong.jin,xiangren\}@usc.edu}\\ {\tt \{maziars,snie,liangtan,mhfirooz\}@fb.com}

}

\maketitle

\begin{abstract}

To reduce a model size but retain performance, we often rely on knowledge distillation (KD) which transfers knowledge from a large ``teacher" model to a smaller ``student" model.
However, KD on multimodal datasets such as vision-language tasks is relatively unexplored, and digesting multimodal information is challenging since different modalities present different types of information.
In this paper, we perform a large-scale empirical study to investigate the importance and effects of each modality in knowledge distillation.
Furthermore, we introduce a multimodal knowledge distillation framework,
modality-specific distillation (MSD), to transfer knowledge from a teacher on multimodal tasks by learning the teacher's behavior within each modality.
The idea aims at mimicking a teacher's modality-specific predictions by introducing auxiliary loss terms for each modality. Furthermore,
because each modality has different saliency for predictions, we define saliency scores for each modality and investigate saliency-based weighting schemes for the auxiliary losses.
We further study a weight learning approach to learn the optimal weights on these loss terms. 
In our empirical analysis, we examine the saliency of each modality in KD, demonstrate the effectiveness of the weighting scheme in MSD, and show that it achieves better performance than KD on four multimodal datasets.



\end{abstract}
\section{Introduction}

\begin{figure}[tb!]
    \centering
    \includegraphics[width=0.85\columnwidth]{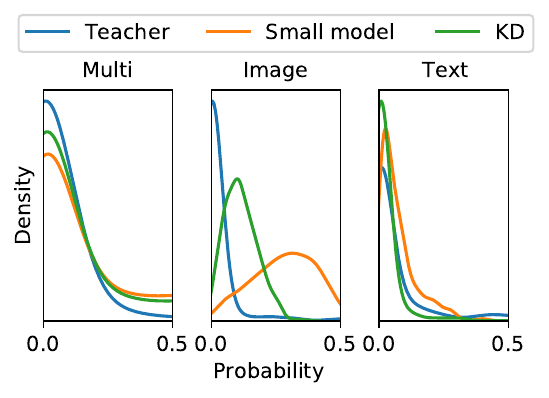}
    \caption{\textbf{Density of model outputs on Hateful-Memes:} given multimodality samples as input (Multi), given only image modality as input (Image), and given only text modality as input (Text). KD denotes a student model with knowledge distillation and the small model is a student model \textit{without} distillation. We observe that there is still a prediction gap between the teacher and the student trained by KD. In this paper, we study saliency explanations for each modality and propose modality-specific distillation (MSD) to minimize the gap. 
}
    \label{fig:example}
\end{figure}

Recent advances in computer vision and natural language processing are attributed to deep neural networks with a large number of layers.
Current state-of-the-art architectures are getting wider and deeper with billions of parameters, e.g., BERT~\cite{Devlin2019BERTPO} and GPT-3~\cite{brown2020language}.
Such wide and deep models suffer from high computational costs and latencies at inference. 
To mitigate the heavy computational cost and the memory requirement, there have been several attempts to compress a larger model (a teacher) into a smaller model (a student)~\cite{ba2014deep,hinton2015distilling,romero2014fitnets,park2019relational,muller2020subclass}.
Among them, \textit{knowledge distillation} (KD)~\cite{hinton2015distilling} assumes the knowledge in the teacher as a learned mapping from inputs to outputs and transfers the knowledge from a larger model to a smaller model.
Recently, KD has been explored in various studies such as improving a student model~\cite{hinton2015distilling,park2019relational,romero2014fitnets,tian2019contrastive,muller2020subclass} and improving a teacher model itself by self-distillation~\cite{xie2020self,kim2020self,Furlanello2018BornAN}.

There has been a lot of interest in multimodal distillation setup such as cross-modal distillation~\cite{gupta2016cross,tian2019contrastive}. 
Multimodal problems involve relating information from multiple sources.
For example, visual question answering (VQA) requires answering questions about an image~\cite{antol2015vqa,goyal2017making,gurari2018vizwiz,singh2019towards} and models should incorporate information from the text and image sources to answer the questions.
Multimodal problems are important because many real-world problems require understanding signals from different modalities to make accurate predictions; information on the web and social media is often represented as textual and visual descriptions.
Digesting such multimodal information in an effective manner is challenging due to their different types of information on each modality.

In this paper, we offer a large-scale, systematic study on the effects of each modality through saliency explanations in KD.
While KD approaches can be applied to multimodal applications, the student and teacher models may significantly differ in their outputs using each modality as input. 
We illustrate the point in Fig.~\ref{fig:example}. 
To minimize the gaps, we introduce a multimodal KD framework, \emph{modality-specific distillation (MSD)}, that aims to mimic the teacher’s modality-specific predictions.


We show that the samples' modalities have a different amount of information.
Based on this observation,
we improve the knowledge transfer by splitting the multimodality into separate modalities, using them as additional inputs, and thus distilling the modality-specific behavior of the teacher.
MSD introduces auxiliary losses per modality to encourage each modality to be distilled effectively.

To maximize the effect of modality-specific distillation, we investigate multiple \textit{weighting schemes} to balance out the auxiliary losses. 
One of the weighting schemes is based on \textit{modality saliency scores}
that are proxy scores to modality importance. 
Furthermore, we leverage a meta-learning method to introduce \textit{weight-learning} to automatically learn optimal weights per sample per modality.



\section{Preliminaries}

In this section, we define notations and revisit conventional knowledge distillation (KD).

\subsection{Problem Definition}
Given a trained and frozen teacher model $T$ and a student model $S$, 
the output of our task is a trained student model.
Our goal is to transfer knowledge from the teacher to the student on multimodal datasets.
We let $f_T$ and $f_S$ be functions of the teacher and the student, respectively. $t$ and $s$ refer to the softmax output of the teacher and the student. Typically the models are deep neural networks and the teacher is deeper than the student. 
The function $f$ can be defined using the output of the last layer of the network (e.g., logits). 
$X$ is a multimodal (language-vision) dataset, $X^\txt$ refers to only the text modality of $X$, $X^\img$ refers to only the image modality of $X$, and $x_i$ is a dataset instance.
In this work, we focus on one text and one image modalities, but it is easy to extend the work to more/other modalities.


\subsection{Conventional Knowledge Distillation}
In knowledge distillation~\cite{hinton2015distilling},
a student is trained to minimize a weighted sum of two different losses: (a) cross entropy with hard labels (one-hot encodings on correct labels) using a standard softmax function, (b) cross entropy with soft labels (probability distribution of labels) produced by a teacher with a temperature higher than 1 in the softmax of both models. The temperature controls the softness of the probability distributions.
Thus, the loss for the student is defined as:
\begin{equation}
    \mathcal{L}_\text{student} = \lambda \mathcal{L}_\text{CE} + (1-\lambda) \mathcal{L}_\text{KD},
\end{equation}
where $\mathcal{L}_\text{CE}$ is a standard cross-entropy loss on hard labels, $\mathcal{L}_\text{KD}$ is a distillation loss, which is a cross-entropy loss on soft labels, and $\lambda \in [0,1]$ controls the balance between hard and soft targets.

To be specific, knowledge distillation~\cite{hinton2015distilling} minimizes the Kullback-Leibler divergence between soft targets from a teacher and probabilities from a student. 
The soft targets (or soft labels) are defined as softmax on outputs of $f_T$ with temperature $\tau$. The distillation loss is defined as follows:
\begin{equation}
    \mathcal{L}_\text{KD} = \tau^2 \frac{1}{|X|} \sum_{x_i \in X} \text{KL} (t(x_i; \tau), s(x_i; \tau)) ),
\end{equation}
where $t(x_i; \tau)=\sigma(\frac{f_T(x_i)}{\tau}), \, s(x_i; \tau)= \sigma(\frac{f_S(x_i)}{\tau})$,
$\sigma$ is a softmax function.
The temperature parameter $\tau$ controls the entropy of the output distribution (higher temperature $\tau$ means higher entropy in the soft labels). Following~\citet{hinton2015distilling}, we scale the loss by $\tau^2$ in order to keep gradient magnitudes approximately constant when changing the temperature.
We omit $\tau$ for brevity.
 



\para{Limitations.}
This KD can be applied to multimodal setups and student models in this distillation are directly trained to mimic a teacher's outputs.
As a result, the student and teacher models may significantly differ in outputs with a single-modality input, i.e., modality-specific outputs, which may lead to inefficient distillation (Fig.~\ref{fig:example}).
To better mimic the teacher's behaviors, we introduce a multimodal KD approach, modality-specific distillation, in the next section.

\section{Analysis Setup}

In this section, we introduce a multimodal KD approach, modality-specific distillation, to understand the importance of each modality (\S\ref{sec:msd}), experimental setup (\S\ref{sec:expsetup}), and datasets for the experiments (\S\ref{sec:dataset}).

\subsection{Modality-specific Distillation}
\label{sec:msd}
The idea of MSD is to feed each modality as a separate input into a teacher and a student, and transfer the modality-specific knowledge of the teacher to the student.
Specifically, MSD introduces two loss terms, $\mathcal{L}_\text{textKD}$ and  $\mathcal{L}_\text{imageKD}$ to minimize difference between probability distributions between the teacher and the student given each modality (assuming text and image as the only two modalities). 
\begin{equation}
    \mathcal{L}_\text{textKD} = \tau^2 \frac{1}{|X_\txt|} \sum_{x_i \in X_\txt} \text{KL} (t(x_i), s(x_i)).
\end{equation}
$\mathcal{L}_\text{imageKD}$ is similarly defined; the input is the image modality instead.

With above two auxiliary losses, the MSD loss for the student is defined as follows:
\begin{multline}
\label{eq:loss}
    \mathcal{L}_\text{MSD} =  \sum_{x_i \in X} w_i \mathcal{L}_\text{KD}(x_i)\\
     + \sum_{x_i \in X^{\img}} w^\img_i \mathcal{L}_\text{imageKD}(x_i) 
    + \sum_{x_i \in X^{\txt}} w^\txt_i \mathcal{L}_\text{textKD}(x_i),
\end{multline}
where we omit the scaling factor $\tau^2 \frac{1}{|X|}$ for brevity. $w_i,w^\txt_i,w^\img_i \in [0,1]$ control the balance between three distillation losses. 
These weights determine the importance of each modality and we hypothesize that the choice of weighting approaches affects the student's performance.
We will introduce four weighting schemes for distillation losses and discuss each of them in \S\ref{sec:weighting}.

\subsection{Experimental Setup}
\label{sec:expsetup}


Through our empirical analysis, we aim to answer the following questions:
\begin{itemize}[leftmargin=*]
\setlength\itemsep{0em}
    \item \textbf{Q1.} How salient is each modality for predictions? 
    \item \textbf{Q2.} Can the saliency explanations aid students?
    \item \textbf{Q3.} Can we learn a sample weighting strategy to better aid students?
    \item \textbf{Q4.} Is the student with the weighting strategies consistent with the teacher?
    \item \textbf{Q5.} Can this be applicable to other distillation methods? 
\end{itemize}

We first define saliency scores for modalities to investigate how salient each modality is for predictions.
(Q1).
Then, we analyze the influence in downstream task performance brought by different weighting schemes for $w_i,w^\txt_i,w^\img_i \in [0,1]$ in MSD (Q2 and Q3). 
For Q4, we examine the student model's sensitiveness to changes in modalities.
Lastly, we try to understand the effect of MSD in various distillation approaches (Q5).


To this end, we use Conventional KD~\cite{hinton2015distilling} as a base distillation approach for MSD. 
In addition, we include several distillation baselines including Conventional KD~\cite{hinton2015distilling}, FitNet~\cite{romero2014fitnets}, RKD~\cite{park2019relational}, and SP~\cite{tung2019similarity} for comparison.
Other distillation approaches are applicable to MSD and we will discuss the results using other KD approaches in our experiments.
To perform analysis, we adopt VisualBERT~\cite{li2019visualbert}, a pre-trained multimodal model, as the teacher model and TinyBERT~\cite{jiao2019tinybert} as a student model. 
VisualBERT consists of 12 layers and a hidden size of 768, and has 109 million parameters, while TinyBERT consists of 4 layers and a hidden size of 312, and has 14.5 million parameters.
We use the region features from images for both the teacher and the student and fine-tune the student on each dataset. 
For training the weight learner we use the datasets' validation set as meta data.
We find the best hyperparameters on the validation set.

\subsection{Datasets and Evaluation Metrics}
\label{sec:dataset}
To answer the questions, we select four multimodal datasets: Hateful-Memes~\cite{kiela2020hateful} MM-IMDB~\cite{arevalo2017gated}, Visual Entailment (SNLI-VE)~\cite{xie2019visual,young2014image}, and VQA2~\cite{goyal2017making}. 

The Hateful-Memes dataset consists of 10K multimodal memes. The task is a binary classification problem, which is to detect hate speech in multimodal memes. We use Accuracy (ACC) and AUC as evaluation metrics for hateful memes.

The MM-IMDB (Multimodal IMDB) dataset consists of 26K movie plot outlines and movie posters. The task involves assigning genres to each movie from a list of 23 genres. This is a multi-label prediction problem, i.e., one movie can have multiple genres and we use Macro F1 and Micro F1 as evaluation metrics following~\cite{arevalo2017gated}.

The goal of Visual Entailment is to predict whether a given image semantically entails an input sentence. Classification accuracy over three classes (``Entailment", ``Neutral" and ``Contradiction") is used to measure model performance.  We use accuracy as an evaluation metric following~\cite{xie2019visual}.

The task of VQA2 is to correctly answer a question given an image. VQA2 is built based on the COCO~\cite{lin2014microsoft} and is split into train (83k images and 444k questions), validation (41k images and 214k questions), and test (81k images and 448k questions) sets. Following the experimental protocol in BUTD~\cite{anderson2018bottom}, we consider it a classification problem and train models to predict the 3,129 most frequent answers. We test models on test-dev of the VQA2 dataset.



\section{Modality Weighting Methods}
\label{sec:weighting}


For the analysis, we introduce three categories of weighting schemes for MSD, presented in the order of complexity: a) population-based (\S\ref{sec:popul}), b) saliency-based (\S\ref{sec:instance}) weighting approaches for the losses, and c) weight-learning approach (\S\ref{sec:meta}) to find the optimal weights. 

\subsection{Population-based Weighting}
\label{sec:popul}

Population-based weighting is to assign weights depending on a modality; we give constant weights $(w_i,w_i^\img,w_i^\txt)$ for each loss term in equation~\eqref{eq:loss}.
This weighting approach assumes the weights are determined by the types of modality. 
Best weights or coefficients for each loss term are obtained by grid search on the validation set.
However, population-based weighting is limited because it does not assign finer-grained weights to each data instance; each data instance might have different optimal weights for the loss terms. This is what we pursue next in saliency-based weighting.

\subsection{Saliency-based Weighting}
\label{sec:instance}
While we observe prediction gaps between the teacher and the student (Fig.~\ref{fig:example})
on each modality, it is unclear which modality leads to such gaps between them and how salient modality is for predictions.
Saliency-based weighting is to give different weights to each loss term depending on a data sample based on its saliency of each modality. 
The assumption is that each data point has different optimal weights for knowledge distillation. 
By assigning instance-level weights, we expect better learning for the student to mimic the teacher's modality-specific behavior.
As it is not possible to tune sample weights as separate hyper-parameters, we instead propose to use simple/intuitive fixed weighting functions, described as follows.  
Obviously, the next step to this approach would be to learn this weighting function alongside the rest of the model, i.e. weight learning, which we discuss further in \S\ref{sec:meta}.

To better understand how these modalities affect the predictions, we first define saliency scores for modalities per sample.
Similar to \citet{li2016understanding}, we erase one of the modalities and measure the saliency score by computing the difference between two probabilities. 
Although the saliency scores can be defined on all inputs, we limit our analysis to explanations to different modalities in this work. 

\para{Quantifying Saliency of Modality.}
Given a teacher model $t$ and a multimodal dataset, we define a saliency score as follows:
\begin{equation}
\label{eq:sal}
    \sal(m) = \delta(t(x), t(x_{-m})), 
\end{equation}
where $m$ denotes a modality and $x_{-m}$ denotes an input after masking out the corresponding modality input. 
$\delta$ is a function to measure difference between $t(x)$ and $t(x_{-m})$. 
We exploit teacher's output to compute saliency scores.
We introduce two saliency-based weighting approaches with different $\delta$ functions.

\para{KL divergence-based weighting.} 
In this weighting approach, $\delta$ is defined as  Kullback–Leibler (KL) divergence which measures the distance between two probability distributions.
Thus, $\delta$ measures distance between predictions with multimodality and predictions by erasing one modality.
Thus, weights for loss terms are defined as $w^\img_i = g (\sal_{i,\txt})$ and $w^\txt_i = g (\sal_{i,\img})$,
where $g =\tanh(\cdot)$ to ensure the weights are in the range $[0,1]$.
In this strategy, we assign $w_i=1$ for the loss term for multimodality. 
Note that in this strategy we do not explicitly use the true labels to decide the distillation weights, and we use the teacher's predictions instead.

\para{Loss-based weighting.} 
Another idea of saliency-based weighting is to weight terms depending on how different loss of predictions with one modality is from loss of predictions with multimodality.
We explicitly use the true labels to measure the loss, i.e., cross-entropy loss. 
If the loss of predictions with one modality is similar to that with multimodality, then we consider the modality salient for predictions.
Thus, the weights are defined as 
\begin{equation}
    w_i: w^\img_i: w^\txt_i = 1: \frac{h(t(x_i))}{h(t(x_i^\img))}: \frac{h(t(x_i))}{h(t(x_i^\txt))},
\end{equation}
where $h(x)= -\sum^c_{j=1} y_{i,j} \log x$ and $y_{i,j} \in \{0,1\}$ are the correct targets for the $j$-th class of the $i$-th example.
In this case, we also assign weights $w_i$ for multimodality depending on the other two weights.  
In order to choose the actual weights, we add a normalization constraint such that, $w_i + w^\img_i + w^\txt_i = 1$. It is worth noting that in this weighting scheme, the actual labels are directly used in deciding the weights unlike the previous one.




\subsection{Weight Learning}
\label{sec:meta}   


Although the aforementioned weighting schemes are intuitive, there is no reason to believe they are the optimal way of getting value out of modality-specific distillation. Moreover, 
it is not trivial to get optimal weight functions since it depends on a dataset.
Thus, we propose a weight-learning approach to find optimal weight functions.
Inspired by \cite{shu2019meta}, we design weight learners to find the optimal coefficients. $(w_i, w^\img_i, w^\txt_i)$ is defined as follows:
\begin{equation}
\label{eq:weight_func}
    (w_i, w^\img_i, w^\txt_i) = f(t(x_i), t(x_i^\img), t(x_i^\txt); \mat{\Theta}) = \mat{w}(\mat{\Theta}), 
\end{equation}
where $\mat{\Theta}$ defines the parameters for the weight learner network, a Multi-Layer Perceptron (MLP) with a sigmoid layer, which approximates a wide range of functions~\cite{csaji2001approximation}.
In general, the function for defining weights can depend on any input from the sample; but here we limit ourselves to the teacher's predictions.

\para{Weight-Learning Objective.}
We assume that we have a small amount of unbiased meta-data set $\{ x_i^{(\text{meta})}, y_i^{(\text{meta})} \}_{i=1}^M$, representing the meta knowledge of ground-truth sample-label distribution, where $M$ is the number of meta samples and $M \ll N$. 
In our setup, we use the validation set as the meta-data set.
The optimal parameter $\mat{\Theta}^*$ can be obtained by minimizing the following cross-entropy loss:
\begin{multline}
    \mathcal{L}_{\text{meta}}(\mat{w}^*(\mat{\Theta})) \\= -\frac{1}{M} \sum_{i=1}^M \sum_{j=1}^c y_{i,j} \log (s(x_i; \mat{w}^*(\mat{\Theta})),
    \label{eq:meta_loss}
\end{multline}
where $\mat{w}^*$ is an optimal student's parameter, which is defined as follows:
\begin{equation}
    \mat{w}^*(\mat{\Theta}) = \argmin_\mat{w} \mathcal{L}_{\text{student}}(\mat{w}, \mat{\Theta}).
\end{equation}
$w^*$ is parameterized by $\Theta$, a weight learner's parameter.

The weight learner is optimized for generating instance weights that minimize the average error of the student over the meta-data set, while the student is trained on the training set with the generated instance weights from the weight learner. 
The algorithm for weight learning is described in \S\ref{append:weightlearning} of appendix.

\section{Empirical Analysis}
\begin{table*}[tb!]
\centering
\caption{\textbf{Main Results.} Mean results ($\pm$std) over five repetitions are reported. MSD outperforms all the KD approaches. Here, we use MSD on top of conventional KD~\cite{hinton2015distilling}. Also, our weight learning for weights shows the best performance.
}
\resizebox{0.98\linewidth}{!}{
\begin{tabular}[t]{L{6cm}|C{2.5cm}C{2.5cm}C{2.5cm}C{2.5cm}C{2.5cm}C{2.5cm}}
\toprule
    \multicolumn{1}{c}{\multirow{2}{*}{\vspace{-2.2mm}\hspace{-20mm}\textbf{Method}}} & \multicolumn{2}{c}{\textbf{Hateful-Memes}} & \multicolumn{2}{c}{\textbf{MM-IMDB}} & \multicolumn{1}{c}{\textbf{SNLI-VE}}  & \multicolumn{1}{c}{\textbf{VQA2 (D)}}  \\ \cmidrule(lr){2-3} \cmidrule(lr){4-5} \cmidrule(lr){6-6} \cmidrule(lr){7-7}
    &\textbf{ACC} &\textbf{AUC}   &\textbf{Macro F1}    &\textbf{Micro F1}   &\textbf{ACC}  &\textbf{ACC}    \\ 
\midrule
Teacher             &65.28   &71.82   &59.92  &66.53   &77.57   &70.91     \\
\midrule
Small model         &60.83 ($\pm$0.20)   &65.54 ($\pm$0.25)  & 38.78 ($\pm$4.03) &  58.10 ($\pm$1.23)  &72.30 ($\pm$0.35)   &64.20 ($\pm$0.56)   \\
Conventional KD~\cite{hinton2015distilling}     &60.84 ($\pm$1.50)   & 66.53 ($\pm$0.27)  & 41.76 ($\pm$4.72) & 58.96 ($\pm$1.62)  &72.61 ($\pm$0.55) &64.70 ($\pm$0.85)   \\
FitNet~\cite{romero2014fitnets}                 &62.00 ($\pm$0.26)  &67.13 ($\pm$0.51)  &46.21 ($\pm$2.12)   &60.46 ($\pm$0.30) &73.06 ($\pm$0.50)  &\textbf{68.08 ($\pm$1.24)}   \\
RKD~\cite{park2019relational}                   &61.43 ($\pm$0.40)  &67.03 ($\pm$0.21)  &51.16 ($\pm$1.64)   &62.52 ($\pm$0.70) &73.09 ($\pm$0.53)  &64.22 ($\pm$0.57)   \\
SP~\cite{tung2019similarity}                    &61.70 ($\pm$1.10)  &66.11 ($\pm$0.45)  &49.07 ($\pm$0.82)   &61.41 ($\pm$0.34) &73.00 ($\pm$0.98)  &64.15 ($\pm$0.81)   \\
\midrule
MSD (Population)          &62.15 ($\pm$1.71)   &67.56 ($\pm$1.21)  &51.85 ($\pm$0.34) &62.13 ($\pm$0.19)  &\textbf{73.64 ($\pm$0.54)}     &64.86 ($\pm$0.63)  \\
MSD (Saliency, KL div)&62.78 ($\pm$1.00)   &67.94 ($\pm$0.52)  &49.20 ($\pm$1.27) &61.84 ($\pm$0.49)  &73.34 ($\pm$0.48)    &64.93 ($\pm$0.48)  \\
MSD (Saliency, Loss)&63.27 ($\pm$0.45)   &67.72 ($\pm$0.82)  &51.02 ($\pm$0.70) &62.05 ($\pm$0.45)  &73.52 ($\pm$0.54) &64.89 ($\pm$0.58)    \\
MSD (Weight learning)       &\textbf{63.86 ($\pm$1.28)}   &\textbf{68.30 ($\pm$0.62)}  &\textbf{53.12 ($\pm$0.08)} &\textbf{63.00 ($\pm$0.09)}  &73.58 ($\pm$0.23)    &64.35 ($\pm$1.56)   \\
\bottomrule
\end{tabular}}
\label{result}
\end{table*}

In this section, we 
revisit and discuss the questions we raised in \S\ref{sec:expsetup}.

\begin{figure}[tb!]
    \centering
    \subfloat[Hateful-Memes]{\includegraphics[width=0.45\columnwidth]{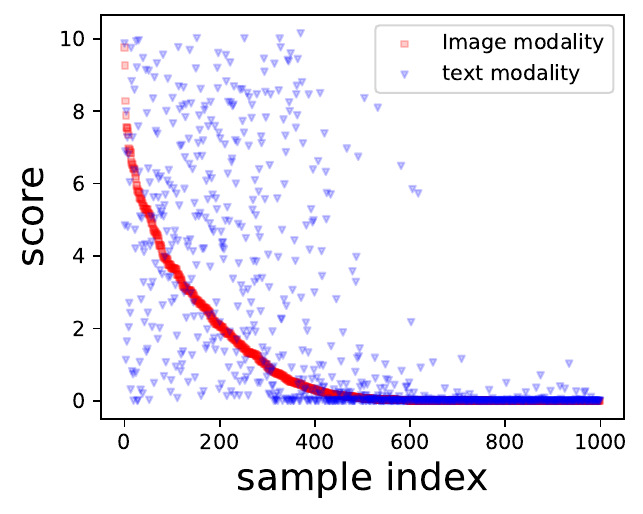}}
    \quad
    \subfloat[MM-IMDB]{\includegraphics[width=0.45\columnwidth]{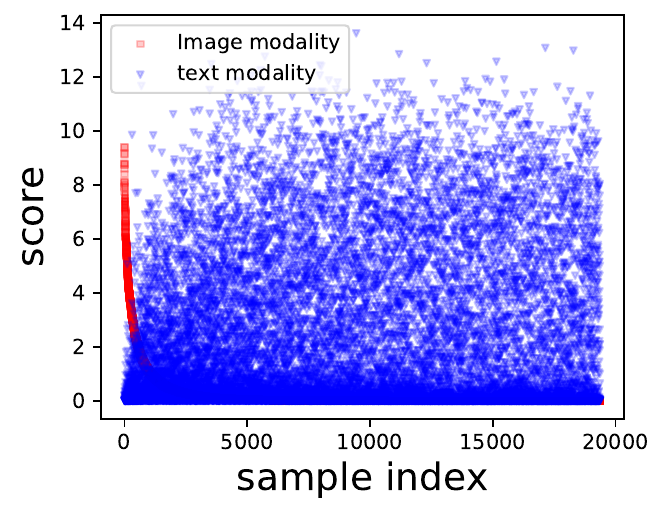} }
    \caption{
    \textbf{Saliency scores in the Hateful-memes and MM-IMDB test sets.} Saliency scores of text modality are mostly higher than those of image modality in MM-IMDB while Hateful-Memes does not show such a global pattern.
    }  
    \label{fig:sal_hmimdb}
\end{figure}

\begin{figure}[tb!]
    \centering
    \subfloat[Entailment]{\includegraphics[width=0.45\columnwidth]{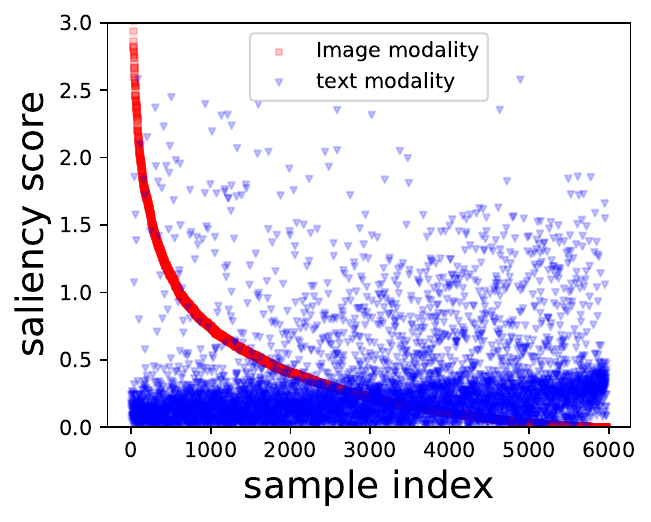}}
    \quad
    \subfloat[Neutral]{\includegraphics[width=0.45\columnwidth]{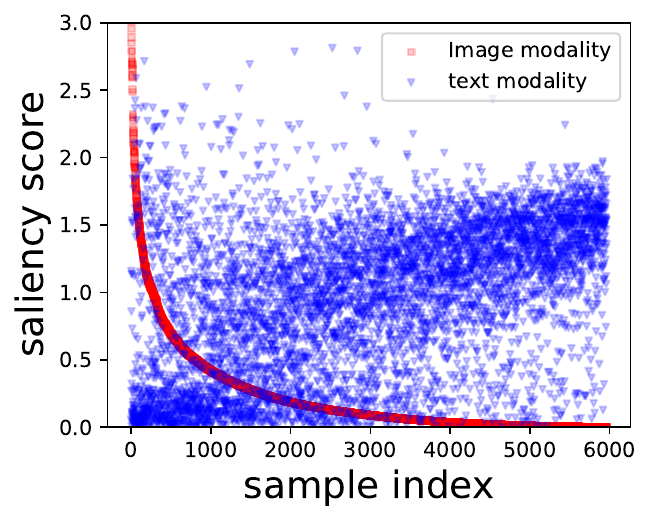} }
    \quad
    \subfloat[Contradiction]{\includegraphics[width=0.45\columnwidth]{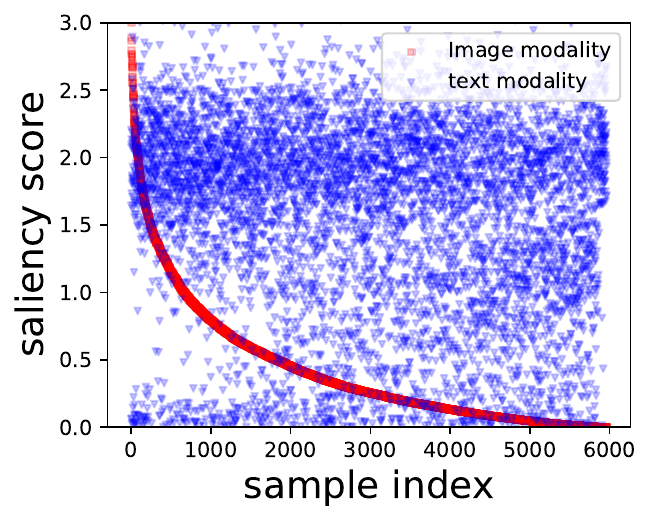} }
    \caption{
    \textbf{Saliency scores in the SNLI-VE dev set.} We observe that saliency scores for text modality are correlated with labels. For the "Entailment" label, scores for text modality are relatively lower, while they are higher for the "Contradiction" label. 
    }  
    \label{fig:sal_vsnli}
\end{figure}


\para{Q1. How salient is each modality for predictions?}
To answer the question, we visualize saliency scores in the Hateful-Memes, MM-IMDB, and SNLI-VE datasets in Figs. \ref{fig:sal_hmimdb} and \ref{fig:sal_vsnli}.
We use KL divergence in Eq.~\eqref{eq:sal}.
We observe that the MM-IMDB dataset shows higher saliency scores of text modality than those of image modality, which implies that text information has important information in general.
On the other hand, Hateful-Memes dataset does not show such a global pattern but one can observe some correlations for individual instances. 
In Fig.~\ref{fig:sal_vsnli}, we notice that saliency scores are correlated with labels in SNLI-VE.
For the "Entailment" label, scores for text modality are relatively lower, while they are higher for the "Contradiction" label, which implies the role of text modality is vital to predict the "Contradiction" label for the teacher model.

\para{Q2. Can the saliency scores aid students?} 
Table~\ref{result} shows our main results on Hateful-Memes, MM-IMDB, SNLI-VE, and VQA2 datasets. 
The small model refers to a student model without knowledge distillation from the teacher. 
As is shown, existing KD approaches improve the student model on all datasets.
However, the MSD approaches improve the small model substantially.
Among them, saliency-based weighting outperforms population-based weighting in the Hateful-Memes dataset.
We note that population-based weighting shows good improvement, which means weighting based on modality only is still very effective on multimodal datasets.
Also, population-based weighting outperforms saliency-based weighting on the MM-IMDB dataset, suggesting all samples are likely to have the same preference or dependency on each modality of the dataset. 
We will discuss results on weight learning in Q3.
Interestingly, FitNet shows the best performance in VQA2. Note that MSD is based on Conventional KD. We will discuss the results of MSD based on other KD approaches in Q5.

\begin{figure}[tb!]
    \centering
    \includegraphics[width=1\columnwidth]{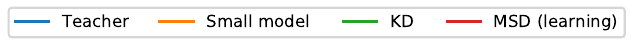}
    {\includegraphics[width=0.85\columnwidth]{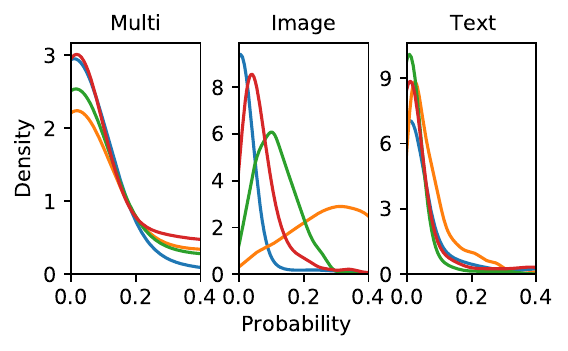}}
    \caption{\textbf{Density of model outputs on samples of label 0 (not hateful) on the test set of Hateful-Memes:} given multimodal samples as input (Multi), given only image modality as input (Image), and given only text modality as input (Text). MSD with the weight-learning approach, minimizes the gap between the teacher and the student trained by KD.
    }
    \label{fig:density_hm}
\end{figure}

\para{Q3. Can we learn a sample weighting strategy to better aid students?}
We observe that among weighting strategies, MSD with weight learning shows the best performance in Hateful-Memes and MM-IMDB, indicating it finds better weights for each dataset in Table~\ref{result}. 
We also find that MSD (Weight learning) shows a similar density curve to the teacher's as shown in  Fig~\ref{fig:density_hm}, which implies that it effectively mimics the teacher's predictions.
However, there is a performance gap between the teacher model and the student model (KD) in predicting true labels given a multimodal sample and each of its individual modalities. 
For example, given only image modality as input (the middle plot in Fig~\ref{fig:density_hm}), there is a considerable difference between the teacher and the small model for predicting benign samples.

\begin{figure}[tb!]
    \centering
    {\includegraphics[width=0.9\columnwidth]{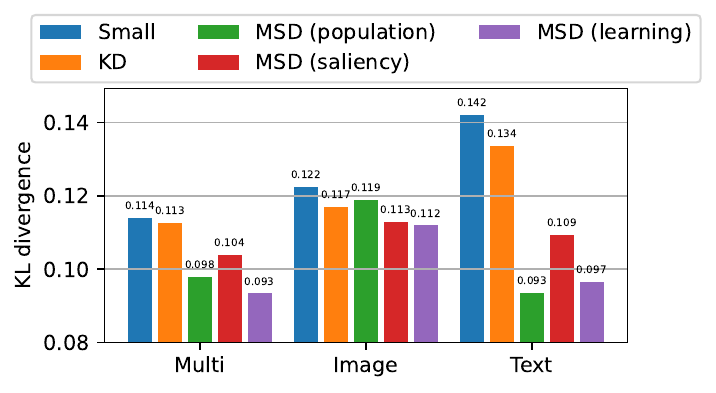}}
    \caption{\textbf{Kullback-Leibler divergence on the MM-IMDB test set between the teacher's outputs and other models' outputs.} This is a measure of how the teacher's probability distribution is different from other models'. The lower divergence is, the closer a model is to the teacher.
    }
    \label{fig:kl_div}
\end{figure}

In addition, we measure Kullback-Leibler (KL) divergence between the teacher's outputs and other models' outputs on the MM-IMDB test set in Fig~\ref{fig:kl_div}.
This is to measure the difference between teacher's probability distribution and others'.
The MSD (learning) approach shows the smallest KL divergence from the teacher which means the student learned with MSD outputs probability distribution close to the teacher's.
Notably, MSD (population) shows the smaller KL divergence than MSD (saliency), which validates that one modality is generally dominant in the MM-IMDB dataset.

\begin{figure}[tb!]
    \centering
    {\includegraphics[width=0.7\columnwidth]{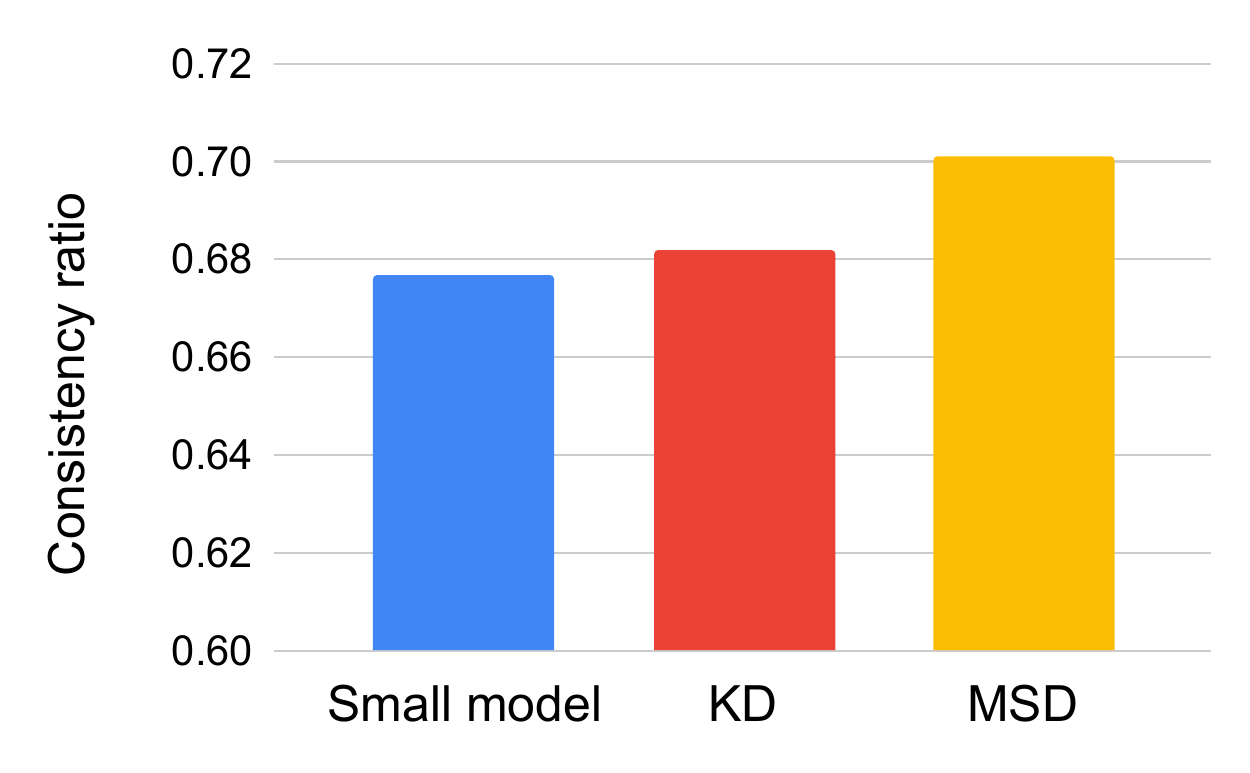}}
    \caption{\textbf{Teacher-Student consistency ratio.} We investigate the student model's sensitiveness to changes in modalities. Higher ratio indicates its sensitiveness is closer to the teacher's.
    }
    \label{fig:consistency}
\end{figure}

\para{Q4. Is the student with the weighting strategies consistent with the teacher?} 
To showcase that our approach helps the student model to be more sensitive to important changes in modalities, we take examples from the Hateful-Memes test set and randomly replace one of the modalities with a modality from another sample. Hateful-Memes is a multimodal dataset and changing the modalities might or might not change the final label. In this case, we do not have the ground truth, but we use the teacher's predicted label on the newly generated sample as a proxy for ground truth and count the times that the student/small model is consistent with the teacher on these generated samples. We define the ratio of such consistent predictions over the total generated samples as ``\textit{Teacher-Student consistency ratio}". Note that none of the models have seen these samples during the training.
As it can be seen from Fig.~\ref{fig:consistency}, the MSD approach with weight learning has a larger ``Teacher-Student consistency ratio" than the small model with and without KD. This indicates that MSD not only improves the accuracy but also improves the sensitivity of the student model to better match the teacher on the changes in modalities on unseen data. 
Please refer to case study in \S\ref{append:case} of appendix.

\begin{table}[tb!]
\centering
\caption{\textbf{Improvement over KD approaches with MSD.} The MSD improves existing KD approaches.
}
\resizebox{0.98\linewidth}{!}{
\begin{tabular}[t]{l|ccccc}
\toprule
    \multicolumn{1}{c}{\multirow{2}{*}{\vspace{-2.2mm}\hspace{-2mm}\textbf{Method}}} & \multicolumn{2}{c}{\textbf{Hateful-Memes}} & \multicolumn{2}{c}{\textbf{MM-IMDB}}    &\multicolumn{1}{c}{\textbf{VQA2}}  \\ \cmidrule(lr){2-3} \cmidrule(lr){4-5} \cmidrule(lr){6-6} 
    &\textbf{ACC} &\textbf{AUC}   &\textbf{Macro F1}    &\textbf{Micro F1}  &\textbf{ACC}    \\ 
\midrule
KD~\cite{hinton2015distilling}     &60.84  & 66.53  & 41.76  & 58.96    &64.70     \\
~+MSD                       &\textbf{62.15}   &\textbf{67.56}   &\textbf{51.85}   &\textbf{62.13}   &\textbf{64.86}   \\
\midrule
FitNet~\cite{romero2014fitnets}     &62.00  & 67.13   &46.21   &60.46    &68.08     \\
~+MSD                       &\textbf{62.22}  &\textbf{68.91}   &\textbf{50.42}   &\textbf{61.43}    &\textbf{68.17}    \\
\midrule
RKD~\cite{park2019relational}     &61.43& \textbf{67.03} & 51.16  &  62.52    &64.22      \\
~+MSD                       &\textbf{62.30}  &66.71   &\textbf{52.56}   &\textbf{63.27}    &\textbf{64.40}    \\
\midrule
SP~\cite{tung2019similarity}    &61.70  &66.11   &49.07   &61.41    &64.15     \\
~+MSD                       &\textbf{62.80}  &\textbf{67.30}   &\textbf{53.29}   &\textbf{63.21}    &\textbf{64.28}    \\

\bottomrule
\end{tabular}}
\label{result_improve}
\end{table}

\para{Q5. Can this be applicable to other distillation methods? }
We present improvements over KD approaches with/without MSD in Table~\ref{result_improve}. We choose the population-based weighting approach in this experiment.
Here, we use MSD on top of each KD approach. Note that the MSD approach is orthogonal to existing KD approaches.
The results show the benefits of the MSD method on top of other approaches; MSD improves diverse KD methods on multimodal datasets. 
Notably, MSD based on FitNet also improves the accuracy on the VQA2 dataset.

\section{Related Work}
\para{Knowledge Distillation.}
There have been several studies of transferring knowledge from one model to another~\cite{ba2014deep,hinton2015distilling,romero2014fitnets,park2019relational,muller2020subclass,tian2019contrastive,Furlanello2018BornAN,kim2020self}.
Ba and Caruana~\cite{ba2014deep} improve the accuracy of a shallow neural network by training it to mimic a deep neural network by penalizing the difference of logits between the two networks.
Hinton et al.~\cite{hinton2015distilling} introduced knowledge distillation (KD) that trains a student model with the objective of matching the softmax distribution of a teacher model at the output layer.
Park et al.~\cite{park2019relational} focused on mutual relations of data examples instead and they proposed relational knowledge distillation.
Tian et al.~\cite{tian2019contrastive} proposed to distill from the penultimate layer using a contrastive loss for cross-modal transfer. 
A few recent papers~\cite{Furlanello2018BornAN,kim2020self} have shown that distilling a teacher model into a student model of identical architecture, i.e., self-distillation, can improve the student over the teacher.

\para{Learning for Sample Weighting.}
Recently, some methods were proposed to learn an adaptive weighting scheme from data to make the learning more automatic and reliable including Meta-Weight-Net~\cite{shu2019meta}, learning to reweight~\cite{ren2018learning}, FWL~\cite{dehghani2017fidelity}, MentorNet~\cite{jiang2018mentornet}, and learning to teach~\cite{fan2018learning,wu2018learning,fan2020learning}.
These approaches were proposed to deal with noisy and corrupted labels and learn optimal functions from clean datasets. 
They are different in that they adopt different weight functions such as a multilayer perceptron~\cite{shu2019meta}, Bayesian function approximator~\cite{dehghani2017fidelity}, and a bidirectional LSTM~\cite{jiang2018mentornet}; and they take different inputs such as loss values and sample features. In our case, we adopt these ideas of meta-learning, specifically Meta-Weight-Net, and utilize it in a different context, i.e. multimodal knowledge distillation.

\para{Bias in Multimodal Datasets.}
Different multimodal datasets were proposed to study whether a model uses a single modality's features and the implications for its generalization properties~\cite{agrawal2018don}.
Different approaches were proposed to deal with such problems where the model overfits to a single modality.
Wang et al.~\cite{wang2020makes} suggest regularizing the overfitting behavior to different modalities.
REPAIR~\cite{li2019repair} prevents a model from dataset biases by re-sampling the training data.
Cadene et al.~\cite{cadene2019rubi} proposed RUBi that uses a bias-only branch in addition to a base model during training to overcome language priors.
In our study, although we do not directly deal with the overfitting phenomena, we use different weighting schemes to better transfer the modality-specific information from the teacher to the student.

\section{Conclusion}
We studied knowledge distillation on multimodal datasets; we observed that conventional KD may lead to inefficient distillation since a student model does not fully mimic a teacher's modality-specific predictions.
To better understand knowledge from a teacher on the multimodal datasets, we introduced saliency scores for a modality and modality-specific distillation; the student mimics the teacher's outputs on each modality based on saliency scores.
Furthermore, we investigated weighting approaches, population-based and saliency-based weighting schemes, and a weight-learning approach for weighting the auxiliary losses to take the importance of each modality into consideration.
We empirically showed that we can improve the student's performance with modality-specific distillation compared to conventional distillation.
More importantly, we observe choosing the right weighting approach boosted the student's performance.
We believe that future work can expand on our methods, and search the space of weighting approaches beyond multimodal setups.

\section*{Acknowledgements}
This research is supported in part by the Office of the Director of National Intelligence (ODNI), Intelligence Advanced Research Projects Activity (IARPA), via Contract No. 2019-19051600007, the DARPA MCS program under Contract No. N660011924033, the Defense Advanced Research Projects Agency with award W911NF-19-20271, NSF IIS 2048211, NSF SMA 1829268, and gift awards from Google, Amazon, JP Morgan and Sony. The views and conclusions contained herein are those of the authors and should not be interpreted as necessarily representing the official policies, either expressed or implied, of ODNI, IARPA, or the U.S. Government. We would like to thank all collaborators in USC INK lab and the reviewers for their constructive feedback that help us improve the paper.

\bibliographystyle{acl_natbib}
\bibliography{bibtex}

\clearpage
\appendix

\section{Weight Learning Algorithm}
\label{append:weightlearning}
\begin{algorithm}[t]
\begin{small}
    \caption{Weight-Learning Algorithm}\label{alg:meta}
    \KwInput{Training data $\mathcal{D}$, Meta-data set $\hat{\mathcal{D}}$, batch size $n,m$, learning rates $\alpha, \beta$, max iterations $T$.}
    \For{$t\gets0$ \KwTo $T-1$}{
        $\{ x, y \} \leftarrow \text{SampleMiniBatch}(D, n)$. \label{alg:batch}\\
        $\{ x^{(\text{meta})}, y^{(\text{meta})} \} \leftarrow \text{SampleMiniBatch}(\hat{D}, m)$.\label{alg:batch_meta}\\
        $\mat{\hat{w}}^{(t)}(\mat{\Theta}^{(t)}) \leftarrow \mat{w}^{(t)} - \alpha \frac{1}{n} \sum_{i=1}^{n} \nabla_{\mat{w}} \mathcal{L}_{\text{student}}(\mat{w}^{(t)}, \mat{\Theta}^{(t)})$ \label{alg:student_pre}\\
        $\mat{\Theta}^{(t+1)} \leftarrow \mat{\Theta}^{(t)} - \beta \frac{1}{m} \sum_{i=1}^{m} \nabla_{\mat{\Theta}} \mathcal{L}_{\text{meta}}(\mat{\hat{w}}^{(t)}( \mat{\Theta}^{(t)}))$ \label{alg:meta_update}\\
        $\mat{{w}}^{(t+1)} \leftarrow \mat{w}^{(t)} - \alpha \frac{1}{n} \sum_{i=1}^{n} \nabla_{\mat{w}} \mathcal{L}_{\text{student}}(\mat{w}^{(t)}, \mat{\Theta}^{(t+1)})$\label{alg:model_update}\\
    }
    \Return Network parameters $\mat{w}^{(T)}, \mat{\Theta}^{(T)}$
\end{small}
\end{algorithm}


Finding the optimal $\Theta^*$ and $w^*$ requires two nested loops; one gradient update of a weight learner requires a trained student on the training set. 
Thus, we adopt an online strategy following \cite{shu2019meta}, which updates the weight learner with only one gradient update of the student.
Algorithm~\ref{alg:meta} illustrates its learning process.
First, we sample mini batches from the training and meta-data sets, respectively (lines~\ref{alg:batch} and \ref{alg:batch_meta}).
Then, we update the student's parameter along the descent direction of the student's loss on a mini-batch training data (line~\ref{alg:student_pre}). 
Note that the student's parameter is parameterized by the weight learner's parameter.
With the updated parameter, the weight leaner can be updated by moving the current parameter $\Theta(t)$ along the objective gradient of equation~\eqref{eq:meta_loss} on a mini-batch meta data (line~\ref{alg:meta_update}).
After updating the weight-learner, the student's parameter can be updated on a mini-batch training data (line~\ref{alg:model_update}). 

\section{Observation of Teacher's Predictions}
\begin{figure}[tb!]
    \centering
    \subfloat[Hateful-Memes]{\includegraphics[width=0.45\columnwidth]{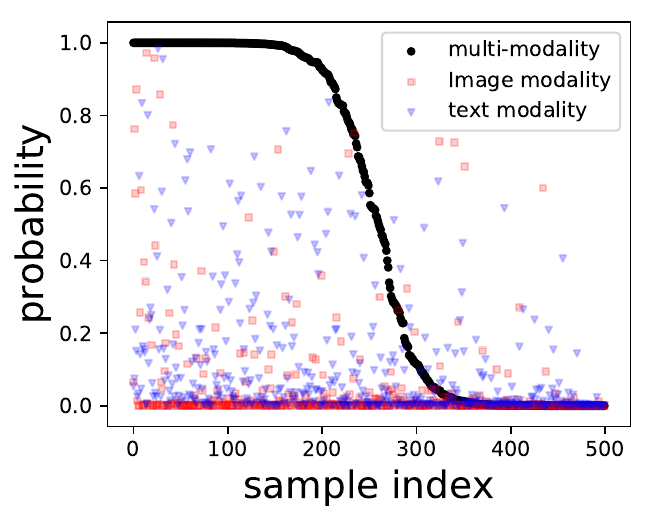}}
    \quad
    \subfloat[MM-IMDB]{\includegraphics[width=0.47\columnwidth]{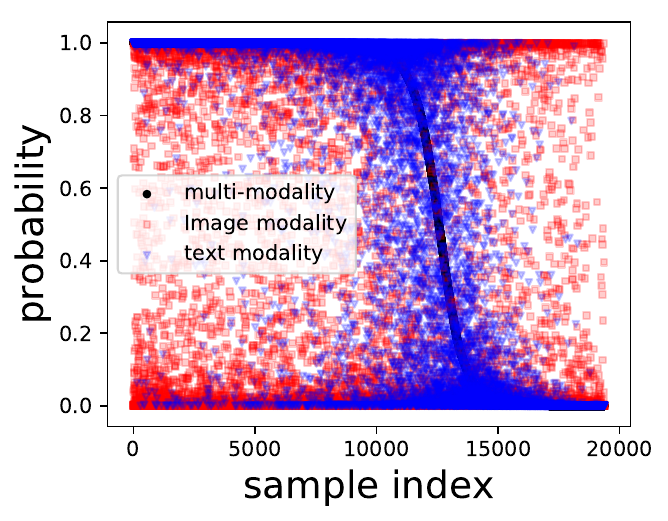} }
    \caption{
    \textbf{Prediction probabilities of test samples for different modalities.} Black points correspond to the predictions of samples with both modalities (original input), red points do with image modality, and blue points do with text modality. The samples are ordered based on their multimodal output probabilities. There is a strong correlation between multimodal predictions and predictions from text modality in MM-IMDB, while there is no such a global pattern in Hateful-Memes.
    }  
    \label{fig:sample_prob}
\end{figure}

Samples from multimodal datasets have different information on each modality. Fig.~\ref{fig:sample_prob} shows a teacher model's predictions for samples in Hateful-Memes and MM-IMDB test sets. 
For each sample, three probabilities are calculated: 1) predictions of samples with both of its modalities, 2) predictions of samples with just its text modality, and 3) predictions of samples with just its image modality. 
As one can see for MM-IMDB there is a strong correlation between multimodal predictions and predictions from text modality, indicating the fact that in MM-IMDB text is a dominant modality. 
On the other hand, for Hateful-Memes dataset there is no such a global pattern but one can observe some correlations for individual instances. 
This behavior is actually expected based on the way Hateful-Memes is built to include unimodal confounders~\cite{kiela2020hateful}. 
Following these observations we introduce four weighting schemes for distillation losses and discuss each of them in \S\ref{sec:weighting}.

\begin{figure}[tb!]
    \centering
    {
        {\includegraphics[width=0.9\columnwidth]{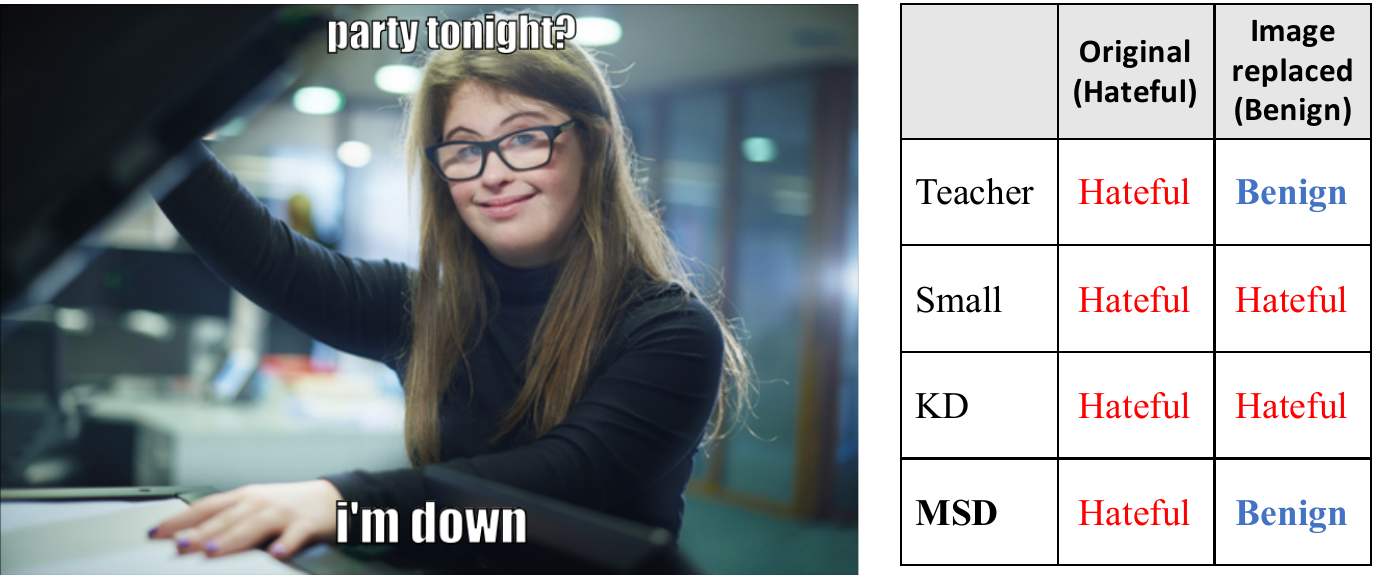}}
        \caption{A multimodal violating sample (Left). We  further replaced its image modality with a background picture that makes it benign and examined models on both examples (Right).
    }
    \label{fig:case}
      }  
\end{figure}

\section{Case Study}
\label{append:case}
We demonstrate the motivation behind our work through an example. Fig.~\ref{fig:case} shows an example of a multimodal sample from Hateful Memes test dataset. The sample is violating based on both modalities together, and all models correctly predict that. To further probe the models, we replace the background image of the sample with a picture that makes the label benign. On this artificially generated sample we notice that only the teacher and MSD model correctly predict benign, while the other two models make wrong predictions (presumably by just looking at the text only). 

\section{Hyperparameters}
The teacher model is a VisualBERT~\cite{li2019visualbert}, and the student model is TinyBERT~\cite{jiao2019tinybert}.
We used the MMF library and pretrained checkpoints from it for VisualBERT\footnote{\url{https://mmf.sh}} and used a pretrained checkpoint in TinyBERT\footnote{\url{https://github.com/huawei-noah/Pretrained-Language-Model/tree/master/TinyBERT}}. 
VisualBERT consists of 12 layers and a hidden size of 768, and has 109 million number of parameters, while TinyBERT consists of 4 layers and a hidden size of 312, and has 14.5 million number of parameters. 
For all experiments, we performed a grid search to find the best hyperparameters.
We adopt the AdamW optimizer to train networks.
We use a linear learning rate schedule that drops to 0 at the end of training with warmup steps of 10\% maximum iterations. 

\para{Hateful-Memes.}
We performed a grid search over learning rates (1e-5, 3e-5, \textbf{5e-5}, 1e-4), and temperatures (1, 2, \textbf{4}, 8), and, batch sizes (\textbf{10}, 20, 30, 40, 50, 60), and the weight learner's learning rates (1e-1, 1e-2, \textbf{1e-3}, 1e-4).
We set the maximum number of iterations to 5000.
The balance parameter $\lambda$ between cross entropy and distillation is set among (0.2, 0.4, \textbf{0.5}, 0.6, 0.8).

\para{MM-IMDB.}
For MM-IMDB experiments, we follow a similar procedure, a grid search, to the Hateful-Memes.
The batch size is 20, temperature is 1, and the weight learner's learning rate is 1e-4.
We set the maximum number of iterations to 10000.
The balance parameter $\lambda$ is set to 0.5.

\para{SNLI-VE.}
For Visual Entailment (SNLI-VE), the batch size is 64, temperature is 4, the student model's learning rate is 1e-4, and the weight learner's learning rate is 1e-4.
We set the maximum number of iterations to 60000.
The balance parameter $\lambda$ is set to 0.6.

\para{VQA2.}
For VQA2, the batch size is 120, temperature is 1, the student model's learning rate is 1e-4, and the weight learner's learning rate is 1e-4.
We set the maximum number of iterations to 60000.
The balance parameter $\lambda$ is set to 0.8.

\begin{table}[t]
    \centering
    \caption{\textbf{Dataset Statistics.}}
    \resizebox{0.9\linewidth}{!}{
    \begin{tabular}[t]{R{2cm}|C{1.5cm}C{1.5cm}C{1.5cm}C{1.6cm}}
         \toprule
         \textbf{Stat.} $\setminus$ \textbf{Data} & \textbf{Hateful-Memes} & \textbf{MM-IMDB} & \textbf{SNLI-VE}    & \textbf{VQA2} \\
         \midrule
         \textbf{Type}   &  Binary & Multilabel  & Multiclass   & Multiclass   \\
         \textbf{\# Classes}    & 2  & 23  &  3 & 3,129 \\
         \midrule
         \textbf{\# Examples}   & 10,000 & 25,959 & 565,286 & 1,105,904    \\
         \midrule
         \textbf{\# Training} & 8,500 & 15,552  & 529,527   & 443,757 \\
         \textbf{\# Validation} & 500 & 2,608  & 17,858 & 214,354 \\
         \textbf{\# Test} & 1,000 & 7,799  & 17,901 & 447,793 \\
         \bottomrule
    \end{tabular}}
    \label{tab:dataset}
\end{table}

\section{Learning Curve}

\begin{figure}[tb!]
    \centering
        {\includegraphics[width=1.0\columnwidth]{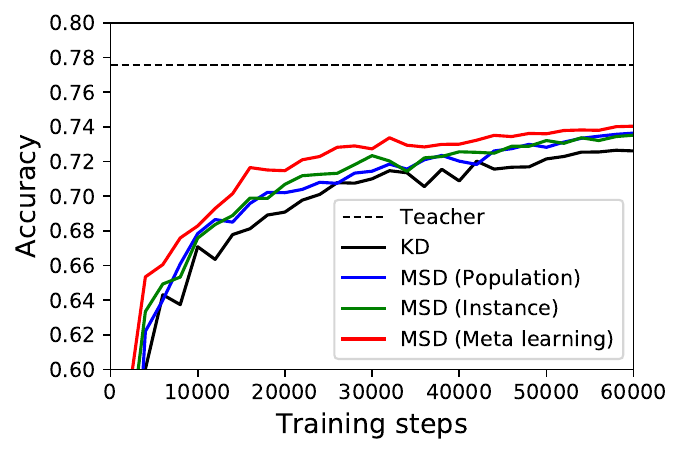}}
        \caption{Test accuracy of a student on SNLI-VE during training and comparison between knowledge distillation (KD) and modality-specific distillation (MSD) with population-based weighting, instance-wise weighting, and weight learning for weights. 
        }
        \label{fig:learning_curve}
\end{figure}

The MSD approaches can also help with training speed, measured by test metrics over training steps.
Fig~\ref{fig:learning_curve} shows the evolution of accuracy on the \emph{test set} during training on the SNLI-VE dataset.
When we train the student with MSD, training progresses faster than KD.
Since the teacher provides two additional probabilities with each modality, the student learns faster and the final performance is better than KD.
We observe a large performance increase early in training with the weight-learning approach, thus leading to the best accuracy.
In this case, the weight learning for sample weighting finds the optimal weights for each data instance, so the student quickly learns from more important modality that is vital for the predictions.

\end{document}